# SiTGRU: Single-Tunnelled Gated Recurrent Unit for Abnormality Detection

Habtamu Fanta[a], Zhiwen Shao[b,c,a,*] and Lizhuang Ma[a,d,*]

[a]*Department of Computer Science and Engineering, Shanghai Jiao Tong University, Shanghai, China*
[b]*School of Computer Science and Engineering, Southeast University, Nanjing, China*
[c]*Key Laboratory of Computer Network and Information Integration (Southeast University), Ministry of Education, Nanjing, China*
[d]*School of Computer Science and Technology, East China Normal University, Shanghai, China*

## ARTICLE INFO

## ABSTRACT

Abnormality detection is a challenging task due to the dependence on a specific context and the unconstrained variability of practical scenarios. In recent years, it has benefited from the powerful features learnt by deep neural networks, and handcrafted features specialized for abnormality detectors. However, these approaches with large complexity still have limitations in handling long-term sequential data (e.g., videos), and their learnt features do not thoroughly capture useful information. Recurrent Neural Networks (RNNs) have been shown to be capable of robustly dealing with temporal data in long-term sequences. In this paper, we propose a novel version of Gated Recurrent Unit (GRU), called Single-Tunnelled GRU for abnormality detection. Particularly, the Single-Tunnelled GRU discards the heavy-weighted reset gate from GRU cells that overlooks the importance of past content by only favouring current input to obtain an optimized single-gated-cell model. Moreover, we substitute the hyperbolic tangent activation in standard GRUs with sigmoid activation, as the former suffers from performance loss in deeper networks. Empirical results show that our proposed optimized-GRU model outperforms standard GRU and Long Short-Term Memory (LSTM) networks on most metrics for detection and generalization tasks on CUHK Avenue and UCSD datasets. The model is also computationally efficient with reduced training and testing time over standard RNNs.

## 1. Introduction

Abnormal Event Detection (AED) is a computer vision task for detecting abnormal events or behaviours from a given image or video. Automatic abnormal event detection has recently gained remarkable attention in the field of computer vision [12, 17], in which humans are freed from controlling surveillance whenever an alert pops up. The surveillance system only focuses on abnormal events while the captured videos contain immense invaluable information. To reduce the unnecessary overheads of reviewing irrelevant videos, it is meaningful to conduct abnormal event detection. However, modelling normal and abnormal video data is a cumbersome task, due to the high-dimensional characteristic of videos, the presence of noises, and the entanglement of various events [8, 22].

In recent years, many abnormal event detection methods are proposed and have achieved great successes [12, 17, 34]. However, it is still a challenging problem due to two key factors. (i) The lack of large-scale annotated training data limits the performance of Deep Convolutional Neural Networks (DCNNs) [28, 29]. (ii) The contentious definition of the term "abnormal" or "anomaly" causes significantly different solutions in different context (environment) [12, 28, 29]. Abnormality is not clearly defined as it is context-specific, leaving a room for unintentional subjectivity in data annotation [12, 17, 28]. For instance, Sun et al. [34] defined "anomaly" as an event or scene that is rarely manifested. Thus, the generalization of abnormal event detection, in which a model built for a particular dataset can detect anomalies from another unrelated dataset, is a tricky and challenging task, as the context on which a system is built plays a big part for its success in another context.

In addition to these main challenges, addressing temporal contexts and capturing long-term dependencies amongst consecutive frames in video data also play a role on the difficulty of abnormality detection tasks [19, 33, 34]. In this paper, we try to address the problem of capturing appearance and motion features in long-term sequences by introducing a new optimized Gated Recurrent Unit (GRU) model that also yields enhanced performance for abnormality generalization task.

In the past few years, most efforts for abnormal event detection resort to complicated handcrafted features. The use of such features has shown limitations as prior knowledge about the context has to be obtained, which is expensive for video scenes [12]. Besides, these handcrafted features extracted from low-level patterns have limited representation capabilities, which are hard to capture motion patterns of complex videos [29]. To alleviate these shortcomings, deep learning based methods have recently taken the stage bringing large performance improvements.

Due to the robust and flexible underlying representation power [17, 28], DCNNs are capable of classifying or regressing concepts by modelling them with large-scale positive samples and can better represent spatial and temporal features. Spatio-temporal based networks have shown their success in learning and representing data with spatial and temporal dimension like videos. Recently, Recurrent Neural Networks (RNNs) have gaining popularity due to their ability of handling sequential data. Generalized RNN and its variant Long Short-Term Memory (LSTM) are widely used for AED and other time sequence tasks such as speech recognition [8, 30]. In addition, over the past years standard RNNs and LSTMs have gone through several structural modifications which led to the rise of a simplified version of LSTM, called GRU. GRUs are widely applied and have shown their effectiveness for speech processing, medical image processing and video segmentation tasks [14, 30, 32]. Despite their consumption for different computer vision problems, abnormal event detection has not taken full advantage of GRUs' inherent nature to develop optimized and effective models, and much effort has not been put forth to exploit GRUs for this task.

*Corresponding authors.
habtamu_fanta@sjtu.edu.cn (H. Fanta); shaozhiwen@seu.edu.cn (Z. Shao); ma-lz@cs.sjtu.edu.cn (L. Ma)
ORCID(s): 0000-0002-8394-8416 (H. Fanta); 0000-0002-9383-8384 (Z. Shao); 0000-0003-1653-4341 (L. Ma)

The main contributions of this work are summarized below:

- We propose a novel, optimized GRU with a single-gate architecture by removing the reset gate from standard GRUs. It can preserve the significant past content without only favouring current input.

- We present an end-to-end deep neural network called Single-Tunnelled GRU (SiTGRU) that achieves computationally efficient and improved abnormality detection performance over standard recurrent neural networks. To our knowledge, this is the first work that employs GRUs for appearance and motion based abnormal event detection task.

- Empirical evaluations show that our method SiTGRU performs effectively better than standard GRU and standard LSTM based methods on AED benchmarks, and outperforms standard recurrent neural networks on generalizing abnormalities across multiple AED benchmarks. Besides, our model performs well on the sole detection of motion abnormalities.

The rest of this paper is structured as follows. Section 2 discusses related works. In Section 3, we explore our methodology and the proposed model structure. Detailed experiments are discussed in Section 4. Our conclusions are drawn in Section 5.

## 2. Related Work

Abnormal event detection in videos has shown remarkable progress in the past few years. Different approaches have been proposed through this period whose underlying assumptions have also shown deviations. A popular way for abnormal event detection comprises two core steps. First, features are learnt only from normal events during training. Then an abnormal event is tested against the "normally" trained model to detect cases that diverge from the normal pattern [17, 29, 34].

### 2.1. Handcrafted Feature Based Method

Hinami et al. [17] proposed an approach that jointly detects and recounts abnormal events by exploiting generic knowledge. Multi-task Fast R-CNN is used to detect distinct visual concepts, which are then supplied to three separate anomaly detectors (viz., One-Class Support Vector Machine, Nearest Neighbour and Kernel Density Estimation) to measure anomaly scores of each visual concept in a scene and identify whether it is anomalous or not. The proposed recounting process provides evidence on the anomalous behavior of a detected abnormal event.

### 2.2. Deep Learning Based Method

**Generative adversarial network based method.** Deep learning through generative adversarial networks (GANs) for abnormal event detection was introduced by Ravanbakhsh et al. [29]. GANs are used to learn and model normal crowd behaviour using unsupervised data. During testing, learnta models are used to produce appearance and motion information. As the network is learnt to generate only normal patterns, it will fail to precisely reconstruct test frames consisting of appearance and motion details of anomalous regions. The difference between the original test frames and the generated frames is used to detect the presence of abnormality in the frame region.

**Spatio-temporal based method.** A spatio-temporal neural network architecture for anomaly detection consisting of two major components was presented by Chong et al. [8], which was evaluated in multiple scenes including crowded scenes. One of the components is responsible for representing spatial features, whereas the other component learns temporal evolution of spatial features. The introduction of autoencoders with LSTMs has shown success in learning temporal patterns and predicting time series information, making them suitable for anomaly detection tasks. The compatibility and efficiency of spatio-temporal autoencoders are further strengthened by introducing 3D convolutional layers [20] in neural networks to learn spatial and temporal features from videos [37]. The typical reconstruction loss in autoencoders is accompanied by introducing a new weight-decreasing prediction loss capable of generating and predicting future frames, hence enhancing learning of motion features. The approach is also typified by the presence of overlapping operations in extracting feature representations. 3D convolution [20] is beneficial for exploiting spatial and temporal details, which indicates that information from multiple channels can be fused together and regularization can be applied on high-level features.

Ravanbakhsh et al. [28] proposed a Binary Quantization layer which uses external hashing method to initialize its convolutional weights, and append it at the end of a CNN. This layer is capable of representing temporal motion patterns in video frames for anomaly detection task. The core concept behind this approach is to show how to easily capture local anomalies by tracking temporal CNN feature variations. Coupling CNN and Convolutional LSTM with autoencoder for the unbounded problem of video anomaly detection was introduced by Luo et al. [26]. The ConvLSTM module is shown to capture appearance changes well and the autoencoder is capable of encoding motion patterns of normal events. However, this method is sensitive to appearance and motion deviations during evaluation. A fully convolutional autoencoder network that generalizes abnormalities across various datasets by learning temporal regularity in videos was presented by Hasan et al. [16]. Conventional handcrafted techniques in Histogram of Oriented Gradients (HOG) [9] and Histogram of Optical Flow (HOF) [3] are used as appearance and motion feature descriptors. The learnt appearance and motion information are then fed to a fully connected autoencoder to model temporal regularity in videos.

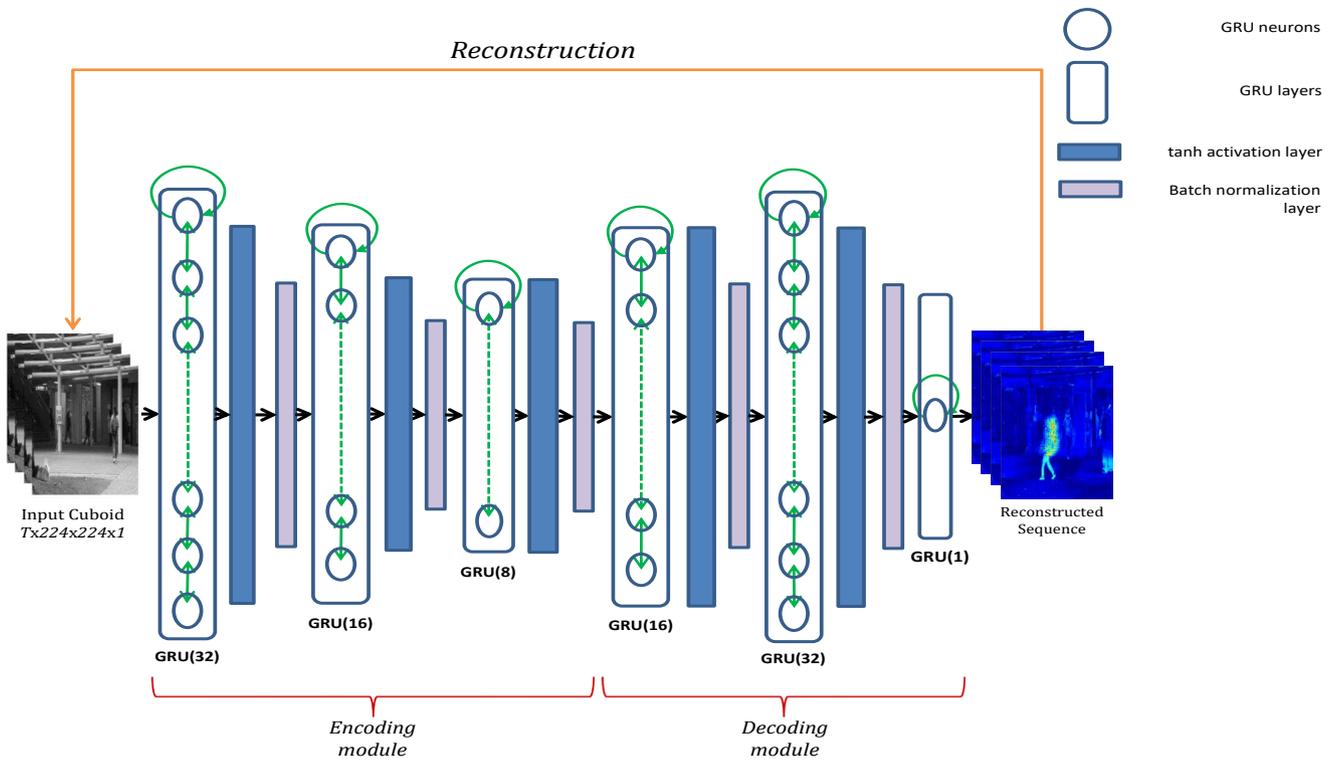

Figure 1: The architecture of our SiTGRU network (appears best in colour and zoom). From left to right: an input layer followed by an encoding module, a decoding module and an output layer. Hyperbolic tangent activation and batch normalization layers consecutively follow each GRU layer (except the last layer). The green arrows inside GRU layers depict intra-connections and recurrent connections in GRU units.

Despite rigorous attempts using handcrafted techniques and variants of deep neural networks, the ability of recurrent neural networks in general and GRUs in particular for abnormal event detection task has been overlooked. Besides, they are not adopted as an integral component for modelling such problems.

## 3. Methodology

Though deep neural networks can achieve excellent performance using a large-scale labelled training data, their ability usually comes short on sequential data [35]. Our proposed Autoencoder-shaped, deep-GRU-based approach analyzes the disparity between consecutive frames to detect the presence of abnormal events. We train an end-to-end model populated with GRU layers that are able to learn long-term sequential patterns and extract spatial features from input video frame sequences. Our trained model reconstructs regular motion features from test videos with little error while incurring higher reconstruction cost for outlier features. As discussed in the related work on abnormal event detection, our model is made to learn normal patterns by training on videos which do not contain any anomalies. To our knowledge, this work is the first attempt to apply GRUs for video abnormality detection and proposes a new, optimized GRU cell that is further extended to produce a novel SiTGRU architecture.

### 3.1. Overview of Our SiTGRU Network

Figure 1 shows the architecture of our SiTGRU network, which learns the spatio-temporal regularity of training videos. It is made up of six GRU layers each consisting of varying number of units. The first and second from last GRU layers contain 32 cells; the second and fourth layers are made of 16 cells while the bottleneck layer in the middle consists of 8 cells. A final GRU layer with a single unit is appended before the output layer to generate a decoded reconstructed sequence with same size as *input cuboid*. Each GRU layer (except the last layer) is followed by hyperbolic tangent activation and batch normalization layers. We apply batch normalization on feed-forward connections to obtain improved system performance and computationally efficient model [18]. Batch normalization lessens internal covariate shift, where distribution of network features varies heavily when training a deep network. It normalizes the activations going into each layer. This enables the mean and variance to be resistant to parametric changes in underlying layers and speeds up the training process. Training GRU networks with batch normalization also generates models that can better generalize themselves to different contexts.

Developing deep RNNs by stacking RNN layers on one another effectively deals with sequence based problems. A stacked GRU based network where a GRU layer is piled on top of another layer with each layer made up of multiple GRU cells, is capable of modelling spatial and temporal information in long sequences. We stack six hidden Single-Tunnelled GRU layers together to build the backbone of the network.

**Input Layer.** We construct the input layer by stacking $T$ consecutive frames in sliding window to make up a temporal cuboid. Building and supplying such cuboid as input to the network enables to better incorporate temporal details of input video frames. This is mainly attributed to the cuboid's ability of keeping appearance and motion patterns, and sustaining temporal information from multiple frames for longer duration [16, 34, 37]. This allows reliable and better feature representation across the model. We do not deploy any feature extraction tasks in the input layer. Instead, we group gray-scale raw frames into cuboid.

**Encoding module.** The Encoding module is structured in a similar way to an encoder module in Autoencoders. Three GRU layers (with 32, 16 and 8 units respectively) each followed by hyperbolic tangent activation and batch normalization layers are stacked together to form this module. These layers are capable of narrowing down the dimensionality of the input cube thereby encoding its features to extract and preserve spatio-temporal information.

**Decoding module.** This module exhibits the structure of an Autoencoder's decoder component. We group two GRU layers (with 16 and 32 units respectively) each appended by hyperbolic tangent activation and batch normalization layers. This module is terminated by a single-unit GRU layer. It rebuilds the encoded sequence fed from encoding module back to the original input. The rebuilt sequence then goes through reconstruction process to generate competitive matching features with input cuboid.

The training process generates a model that stores the network architecture and model weights of an anomaly-free context. During testing, we feed the model with a sequence of frames mostly containing anomalous scenes. The model is expected to be robust enough to effectively identify anomalous events as it is not trained with such content. We employ a reconstruction loss mechanism (discussed in Section 3.4) to evaluate the capability of our trained model in generating the reconstructed sequence back to input test frames. A robust model is expected to carry out the reconstruction process of anomaly-free pixels with minimal cost. If an anomaly exists in a test frame, the model struggles to perfectly reconstruct these pixel regions resulting in larger cost. We perform a series of training procedures by modifying different components and parameters in our network and pick the model that produces the best evaluation with least cost.

### 3.2. Standard GRU

Computer vision problems such as abnormal event detection require handling temporal dependencies among inputs, and modelling short-term and long-term sequences. Recurrent neural networks are capable of handling and processing such kind of sequential data in a better way. In contrast to traditional neural networks, RNNs focus on manipulating state neurons to learn contextual relations in and between sequential data [27]. The limitation that comes with such recurrent networks is the presence of vanishing and exploding gradients that make training RNNs a complex task.

GRUs are able to solve the vanishing and explosion of gradients manifested in conventional recurrent neural networks [5, 6]. The most widely adopted RNNs are LSTM networks that are shown to achieve state-of-the-art performance on various machine learning and deep learning tasks. As a variant of LSTM, GRU is shown to produce equally competent results to LSTM. GRUs differ from LSTMs in that their gates monitor the flow of information from previous time steps while the three gating mechanisms in LSTMs control the flow of information within internal cell unit [13]. GRUs deploy two peculiar gates, update gate and reset gate, to address the vanishing gradient problem of RNNs and dynamically handle temporal information in long-term video sequences. These gating mechanisms allow such RNN variants to be trained to keep necessary and relevant information for longer period from previous states, or discard information with less importance from previous states [5, 6]. Recent works have shown the importance of RNNs with gating mechanisms in achieving improved results for classification and generation tasks with sequence modelling [2, 10, 35]. Recurrent neural networks' inherent capability of adequately modelling sequential data supplemented by the advantages of gated features in GRUs enables them to effectively model tasks that use short-term or long-term video sequences. For such sequences, maintaining temporal features across frames is invaluable for feature learning. Thus, GRUs' gating mechanisms provide a trained model with consistent memory capable of seizing both short-term and long-term dependencies amongst video frames more effectively.

The architecture of GRU is illustrated in Figure 2. GRU is a simplified version of LSTM where only two gates are deployed to generate candidate memory by using current input and previous memory state. As shown in Figure 2, at each time step $t$, a GRU cell takes the contents of previous hidden state $h_{t-1}$ and current input $x_t$, manipulates them through reset and update gates, and sends the computed current state $h_t$ to the next time step. The following formulae define a standard GRU cell:

$$z_t = \delta(W_z x_t + U_z h_{t-1} + b_z), \tag{1a}$$

$$r_t = \delta(W_r x_t + U_r h_{t-1} + b_r), \tag{1b}$$

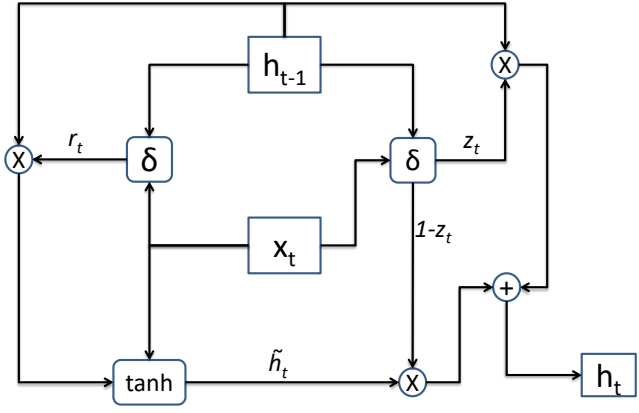

Figure 2: The structure of a standard GRU cell.

$$\tilde{h}_t = \tanh(W_h x_t + U_h(h_{t-1} \otimes r_t) + b_h), \quad (1c)$$

$$h_t = z_t \otimes h_{t-1} + (1 - z_t) \otimes \tilde{h}_t, \quad (1d)$$

where $z_t$ represents the update gate in vector form, $r_t$ denotes vectorized form of the reset gate, and $h_t$ is computed as a linear interpolation of the previous state $h_{t-1}$ and the current candidate memory $\tilde{h}_t$ using the result from update gate. Sigmoid activation $\delta(\cdot)$ is applied on both update and reset gates to squash values between 0 and 1. $\tilde{h}_t$ is computed from a hyperbolic tangent activation $\tanh(\cdot)$. $\otimes$ denotes Hadamard product (element-wise multiplication). $x_t$ is the current input fed into the network; $W_z$, $W_r$ and $W_h$ are trainable weights of feed-forward connections, whereas $U_z$, $U_r$ and $U_h$ are weights of the recurrent connections. $b_z$, $b_r$ and $b_h$ are the bias vectors.

Although standard GRUs are capable of effectively handling long-term sequential data and ease the vanishing or exploding gradient problem of RNNs, their gating structures can lead to the omission of crucial content in a long sequence. Unless their gating structures are monitored, GRUs may lead to poor models where important information from previous time steps and long-term event dependencies are not well addressed during training. In this paper, we present an approach that alleviates this problem by introducing a novel GRU model which is capable of sustaining crucial content in long-term sequential data.

### 3.3. Our Proposed GRU

The major modifications we carry out to conventional GRU cells consist of completely discarding the reset gate and substituting the candidate memory's hyperbolic tangent activation $\tanh(\cdot)$ in Eq. (1c) with the sigmoid activation $\delta(\cdot)$.

#### 3.3.1. Discarding the reset gate

In GRUs, the reset gate in Eq. (1b) decides how relevant the previous memory state $h_{t-1}$ is for computing the candidate memory state $\tilde{h}_t$. It allows the network to either forward the information from previous state or completely leave if deemed unnecessary. The reset gate is ideal for scenarios where considerable interruptions are manifested in sequential data [30]. For such cases, the reset gate can be turned on fully to attain values closer to zero so as to diminish the impact of the previous memory state on computation of candidate memory state.

A scenario of removing or replacing the reset gate in GRUs was applied by Li et al. [24] to selectively read text segments by skipping less important content. A binary input gate is used to replace the reset gate that can easily select more relevant words in a text corpus and effectively model information flow. For abnormal event detection task, memory state of frame sequences which participate in making up a video needs to be maintained when deciding to generate a candidate memory state. Sustaining smooth continuity amongst frames in such a way enables keeping temporal information for longer period, not sacrificing previous memory state $h_{t-1}$ in favour of current input $x_t$. Limiting the participation of previous memory state by applying the reset gate can be dangerous leading to poorly computed candidate state that in turn affects reliability of the current memory state $h_t$. Thus, the reset gate can be discarded for our task as abnormality detection problems are expected to reasonably and unbiasedly treat information from previous and current frame sequences.

Ravanelli et al. [30] assessed the functionality of update and reset gates for speech recognition process, and argue that redundancies might occur in the activations of these two gating mechanisms. Setting similar large or small values to both update and reset gates produces similar effect on candidate memory and current memory states. This initiates the need to discard the reset gate from a GRU cell. Discarding the reset gate from a GRU cell is achieved by removing Eq. (1b) in GRU formulation and modifying Eq. (1c) to the following form:

$$\tilde{h}_t = \tanh(W_h x_t + U_h h_{t-1} + b_h), \quad (2)$$

where abandoning the reset gate in such a way produces computationally lightweight GRU cell and model structure.

#### 3.3.2. Incorporating sigmoid activation

The other modification we carry out on standard gated recurrent unit is replacing candidate memory state's hyperbolic tangent activation with sigmoid function. Doing so alters calculating the candidate memory $\tilde{h}_t$ as the following form:

$$\tilde{h}_t = \delta(W_h x_t + U_h h_{t-1} + b_h). \quad (3)$$

Standard GRUs use hyperbolic tangent function for activation, but these functions are shown to be ineffective for training feed-forward connections as they suffer from performance drop when the network gets deeper [15]. Similarly, Rectified Linear Unit (ReLU) activations portray numerical instabilities emanating from their unbounded nature when used for long range sequences [30]. We thus employ sigmoid activation to compute candidate memory state. Experimental evaluations demonstrate the superiority of models trained with sigmoid activation for abnormality detection task. The modifications we make on a standard GRU cell are shown in

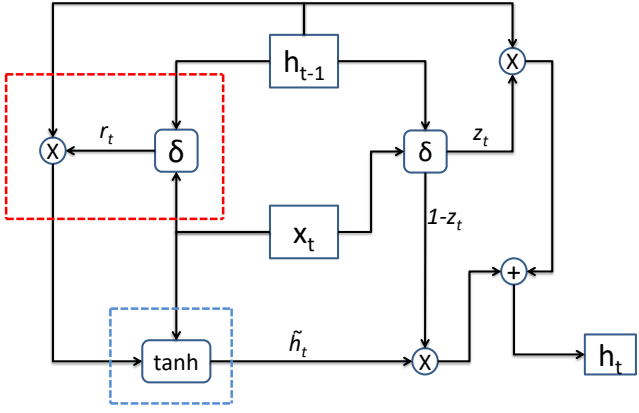

Figure 3: Modifications on the standard GRU cell. The red-dashed box represents the reset gate removed from standard GRU; the blue-dashed box shows standard *hyperbolic tangent* activation replaced with *sigmoid* activation.

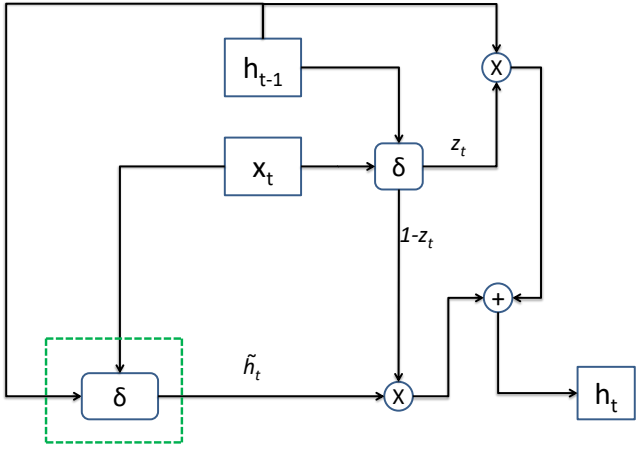

Figure 4: The structure of our proposed GRU cell. The green-dashed box indicates *sigmoid* activation substituting standard GRU's *hyperbolic tangent* activation.

Figure 3, and our proposed GRU cell structure is shown in Figure 4.

The single-tunnelled GRU, where the reset gate is turned off and the standard hyperbolic tangent activation function is substituted with a sigmoid activation, achieves improved detection performance than conventional GRU. The fewer number of parameters obtained by removing $W_r$, $U_r$ and $b_r$ from the reset gate also results in lesser computational cost. Sustaining previous content by allowing it to contribute to computation of candidate memory allows the GRU cell to get diverse features from a wide range of frame sequences. This allows the proposed GRU cell and deep-GRU model to better adapt and generalize themselves to different contexts. Generalization study (discussed in Section 4.5) highlights generalizability performance gains obtained by our proposed approach over conventional recurrent networks on AED benchmarks. Collectively, the proposed changes on standard GRU lead to the following new model:

$$z_t = \delta(W_z x_t + U_z h_{t-1} + b_z), \quad (4a)$$

$$\tilde{h}_t = \delta(W_h x_t + U_h h_{t-1} + b_h), \quad (4b)$$

$$h_t = z_t \otimes h_{t-1} + (1 - z_t) \otimes \tilde{h}_t. \quad (4c)$$

### 3.4. Regularity Score

Regularity score is a widely used metrics to compute the consistency of a test data pattern against a pattern learnt by a trained model [8, 16]. After training, we evaluate model performance by analyzing how well it is capable of detecting abnormal events (unseen during training) by providing with test data containing anomalous frames. The model is expected to detect and discriminate events deviating from the learnt pattern while keeping false alarms at bay. We compute the regularity score of a test video against its reconstruction error. The reconstruction error is calculated as the Euclidean distance between an input test video (cuboid) and its reconstructed frame sequence:

$$r_e(t) = \sqrt{(x(t) - m_W(x(t)))^2}, \quad (5)$$

where $x(t)$ is the $t$th test frame and $m_W$ denotes the learnt model weights of the spatio-temporal deep-GRU model. The overall reconstruction error of an input video is the average reconstruction error of individual input cuboids with a mini-batch size of $N$.

The regularity score $r_s(t)$ of a test video $t$ is calculated using the equation:

$$r_s(t) = 1 - \frac{r_e(t) - \min(r_e(t))}{\max(r_e(t))}, \quad (6)$$

where the expression on the right hand side of the subtraction operation evaluates the anomaly score of a test video using volume reconstruction cost from Eq. (5). It tries to minimize the reconstruction cost by subtracting the least anomalous frame's cost from each frame's cost, and divides it by the highly anomalous frame's reconstruction cost. We apply bilinear interpolation when computing frame reconstruction costs from the obtained volume reconstruction cost.

### 3.5. Anomaly Detection

The process of inspecting and identifying an anomalous frame is accomplished by computing the reconstruction error, a measurement known for its deployment in determining anomalous scenes in test frames. In this paper, we evaluate the anomaly detection accuracy using frame-level criterion. True Positive Rate (TPR) and False Positive Rate (FPR) are obtained by dynamically assigning different error threshold values that are in turn used to compute Area Under Receiver Operating Characteristic Curve (AUC) and Equal Error Rate (EER). AUC by EER is a widely used metrics for performance evaluation in abnormal event detection and related problems. TPR and FPR are calculated as:

$$TPR = \frac{TruePositive}{TruePositive + FalseNegative}, \quad (7a)$$

$$FPR = \frac{FalsePositive}{TrueNegative + FalsePositive}. \quad (7b)$$

### 3.6. Abnormality Generalization

Generalizing abnormalities across different datasets and developing generalized models that can detect abnormalities from multiple datasets has long been a challenge in AED task. We tackle this problem by learning models from a single and multiple dataset.

After training and obtaining models which have learnt spatio-temporal features from a dataset with its specific context, abnormality generalization evaluates the robustness and sensitivity of these models by testing on a completely different dataset. For multiple dataset learning, we build a model by training on a dataset prepared by collecting samples from CUHK Avenue and UCSD datasets. Even though the evaluation context is different from the training context for generalization task (unlike detection task), we use same AUC by EER measure for performance computation on test sets as these tasks are similar except a difference in the approaches pursued.

## 4. Experiments
### 4.1. Datasets and Setting
#### 4.1.1. Datasets

We train and test our model on three famous video anomaly detection datasets: CUHK Avenue [25], UCSD Ped1 and UCSD Ped2 [23]. These datasets are made up of videos taken from a stationary camera in an outdoor environment. Training sets are populated with videos containing only normal events while test sets contain videos with both normal and abnormal scenes (with abnormal scenes having the lion's share). The abnormality of the events varies from one dataset to another; i.e., events that are identified (classified) as abnormal in the test set of Avenue dataset are not the same for UCSD, and vice versa. In all datasets, we use training sets for model learning and determining the structure of the network, whereas test sets are merely used to evaluate performance of trained models.

**CUHK Avenue** is prepared from videos captured at the avenue of Chinese University of Hong Kong campus. It contains 16 training and 21 testing videos with 15,328 frames in train set and 15,324 frames in test set [25]. Each frame has a resolution of $640 \times 360$ pixels. Pixel-level masks are provided for ground truth annotation for frames in test set. The test set has 47 instances of abnormalities manifested over 3,820 anomalous frames. Fourteen events (viz., walking in wrong direction towards the camera, running or romping across a walkway, throwing objects like paper or bag, and pushing a bike) are treated as abnormal behaviour in Avenue dataset.

**UCSD** is made up of videos taken from a pedestrian walkway, and is considered one of the most challenging datasets for abnormal event detection task [23]. It has two subsets, Ped1 and Ped2. Ped1 is composed of 34 training and 36 testing videos whose frames have a resolution of $238 \times 158$ pixels. Ped2 contains 16 training and 12 testing videos with $360 \times 240$ frame resolution. Each test video in Ped1 is sliced into 200 frames, whereas videos in Ped2 test set are sliced into 120, 150 and 180 frames. Ped1 is provided with full frame-level ground truth annotation and pixel-level annotation for 10 videos, whereas Ped2 has frame-level and pixel-level ground truth annotation for all test videos. UCSD treats the presence of car, wheelchair, bicycle, skateboard, and stepping on grass or wrong direction movement across the walkway as abnormal behaviour. Ped1's test set contains a total of 7,200 frames of which 4,005 frames exhibit abnormal behaviour. On the other hand, Ped2's test set comprises of 2,010 frames from which 1,636 frames are treated as anomalous.

#### 4.1.2. Implementation Details

**Preprocessing.** In this stage, raw video data obtained from public datasets are converted into a type that can easily be provided as input to the model. Each raw video in both training and testing set is converted into frames whose size is rescaled to a common $224 \times 224$ pixels. To make sure that these input frames are on a similar scale, we scale down the pixel values to an interval of $[0, 1]$. We then subtract every frame from the global average image to normalize the data and average the pixel values at each location in the training set. The image frames are finally changed to gray-scale and normalized to attain zero mean and unit variance.

The input to the model is a video cuboid constructed from sequential frames with varying number of skipping strides. To cater for the large number of parameters and increase the training set, we adopt a technique followed by Hassan et al. [16] and carry out data augmentation across temporal dimension. Building the input to the model as stack of $T$ frames ($T = 4$ in all our setups) making up a cuboid helps to better entertain and deal with the temporal dimension of videos during training and testing. The constructed input cuboid has a size of $T \times 224 \times 224 \times 1$.

**Training.** We use 85% of the train set on both Avenue and UCSD datasets for training the network and the rest fifteen percent as validation set to validate the model. We train the proposed network for 60 epochs with the mini-batch size $N = 8$. We use the Adam optimizer [21] with a learning rate of 0.00001, $\beta_1 = 0.9$, and $\beta_2 = 0.9999$ for optimizing our end-to-end network. Adam is computationally efficient and inexpensive, capable of handling large data and parameters, and can automatically adjust the learning rate based on history of model weight updates.

We empirically witness that Adam produces models that yield better performance evaluation than AdaGrad (Adaptive Gradient) [11] and SGD (Stochastic Gradient Descent) [31] optimizers. While training on all datasets, we customize different parameters, experiment on a varying number of epochs (to determine the convenient epoch), and pick the model output whose objective function produces the best AUC by EER result on the test set. We experimentally evaluate that sparse categorical cross entropy loss yields best detection results on UCSD Ped1 and Ped2 test sets, whereas mean squared error loss achieves best results on Avenue test set.

### 4.1.3. Evaluation Metrics

We employ the regularity score computed using Eq. (6) to assess the normality of frames in a video and detect the presence of anomalous events. We evaluate our method quantitatively using AUC by EER score, a frequently adopted evaluation metric for such tasks [16, 37]. The ROC curve is generated by plotting the TPR against FPR along varied threshold values. AUC measures the ability of a trained model to precisely distinguish features between multiple classes. Larger AUC scores show better discrimination capability of models to several classes while lower scores depict degraded discrimination ability. EER estimates the error border where FPR and False Negative Rate (FNR) attain proximal values. It is the point on the ROC curve where the curve and the diagonal of unit square intersect [34], i.e., $FPR = 1 - TPR$, as shown in Figure 5. The AUC value signifies the robustness and efficiency of a learnt model to evaluation on unseen data. Following [4], AUC can be computed as the integral of the TPR value by FPR:

$$AUC = \int_0^1 TPR(x)dx, \qquad (8)$$

where $x$ denotes the FPR value.

### 4.2. Comparison with Related Methods

We compare our method SiTGRU with related works including Conv-AE [16], ConvLSTM-AE [26] and STAE-OF [37], and standard RNNs. Figure 5 presents the AUC by EER result of our SiTGRU on Avenue, Ped1, and Ped2 datasets respectively. Table 1 presents results measured with AUC by EER metrics of different methods on Avenue, Ped1 and Ped2 datasets. Evaluation is carried out by training models on a particular dataset's train set and testing with a similar dataset's test set. This table summarizes best test results obtained, and compares performance with standard GRU, standard LSTM, SiTGRU with (w/) ReLU activation and other related works. Our deep-GRU-based approach yields better detection results than standard RNNs on UCSD Ped1 and Ped2 test sets, while obtaining detection performance lesser by a slight margin than standard GRU's performance on Avenue test set.

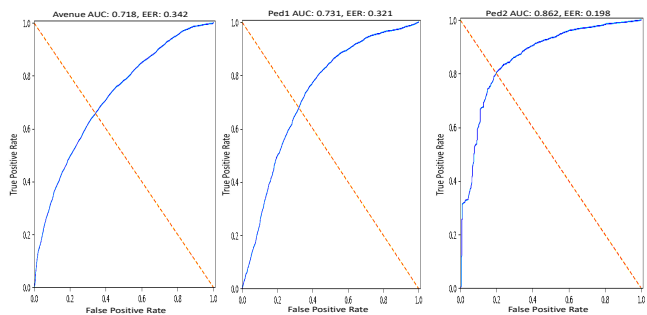

Figure 5: Generated ROC curves of our SiTGRU on Avenue, Ped1 and Ped2 datasets, respectively (appears better in zoom).

| Method | Avenue AUC | Avenue EER | Ped1 AUC | Ped1 EER | Ped2 AUC | Ped2 EER |
|---|---|---|---|---|---|---|
| Conv-AE | 0.702 | 0.251 | 0.810 | 0.279 | 0.900 | 0.217 |
| ConvLSTM-AE | 0.770 | – | 0.755 | – | 0.881 | – |
| STAE-OF | 0.809 | 0.244 | 0.871 | 0.183 | 0.886 | 0.209 |
| GRU | 0.725 | 0.338 | 0.723 | 0.326 | 0.859 | 0.212 |
| LSTM | 0.711 | 0.345 | 0.674 | 0.345 | 0.847 | 0.243 |
| SiTGRU w/ ReLU | 0.641 | 0.400 | 0.712 | 0.334 | 0.708 | 0.327 |
| **SiTGRU** | **0.718** | **0.342** | **0.731** | **0.321** | **0.862** | **0.198** |

Table 1: Performance comparison with standard RNNs and other methods on CUHK Avenue, UCSD Ped1 and UCSD Ped2 datasets.

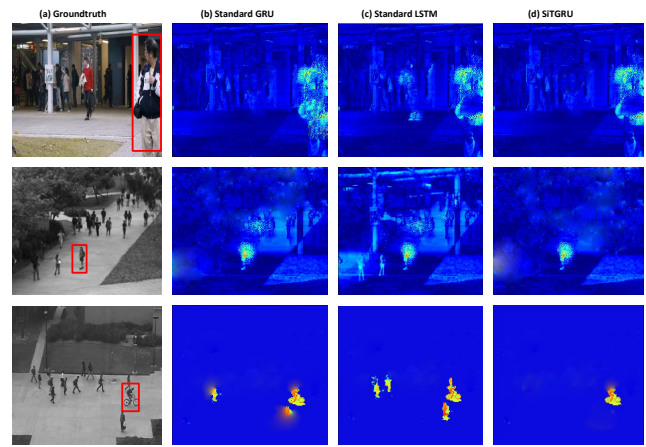

Figure 6: Anomaly visualization of SiTGRU and standard RNNs. Top-to-bottom: **(a)** groundtruth frames from Avenue, Ped1 and Ped2 test sets, **(b)** output heatmap with standard GRU, **(c)** output heatmap with standard LSTM, **(d)** output heatmap with SiTGRU (appears best in colour).

Our SiTGRU performs better than standard LSTM based models on all benchmarks. It attains a 5.7%/2.4% AUC-by-EER improvement on Ped1, 1.5%/4.5% improvement on Ped2, and 0.7%/0.3% improvement on Avenue test sets. Compared to standard GRU, it also obtains AUC-by-EER gains of 0.8%/0.5% and 0.3%/1.4% on Ped1 and Ped2 test sets, respectively. There is little drop in detection performance on Avenue test set which arises from the presence of camera shakes and occlusion of anomalous events (appearance and motion) in few videos.

An abnormality that appears in a particular frame (which is subject to noise or occlusion) loses its previous content whenever a noise prevails or whenever its appearance is obstructed by another object. Standard GRUs are capable of dealing with such disturbances through the reset gate that prioritizes content at current time step limiting participation of previous noisy, occluded content from computation of candidate memory state. This gives them slight edge during model learning and hence better abnormality discrimination. Our SiTGRU still performs better than standard LSTMs and SiTGRU w/ ReLU on such cumbersome scenarios. This shows that if coupled with noise reduction and occlusion-resistant features, the performance of our SiTGRU can be enhanced well for different environments.

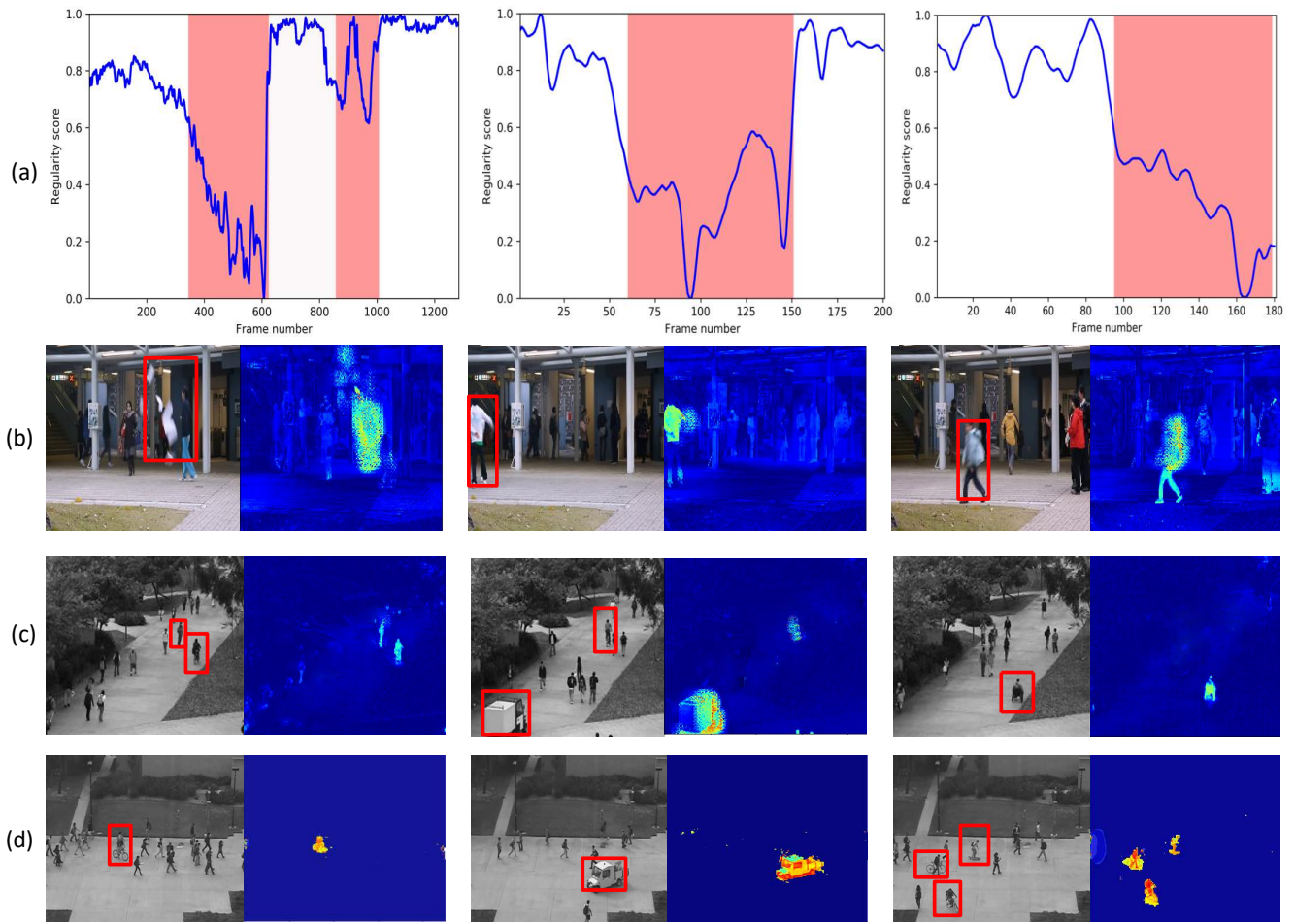

Figure 7: Left-to-right: **(a)** sample regularity score of test videos from Avenue, Ped1 and Ped2 datasets respectively; **(b)** pair of frames with anomalous events (red rectangle) and generated sample frame heat maps from Avenue dataset (left to right: throwing, sporting and romping respectively); **(c)** pair of frames with anomalous events (red rectangle) and generated sample heat maps from Ped1 dataset (left to right: skating and biking, car, and wheelchair respectively); **(d)** pair of frames with anomalous events (red rectangle) and generated sample frame heat maps from Ped2 (left to right: biking, car, and skating and biking respectively) dataset. On the heat maps, regions highlighted in dark yellow and red show pixels that the model failed to reconstruct because of the presence of anomalies, whereas blue regions denote normal, anomaly-free pixels (appears best in colour).

### 4.3. Qualitative Analysis

After model training, the reconstruction error is used to compute regularity score of test videos. Regular frame sequences show lower reconstruction error while anomalous sequences manifest higher error. We qualitatively evaluate the performance of our model with test datasets in Figures 6 and 7. The qualitative evaluation asserts the reported quantitative results in Table 1 and correlates frame behaviour with detected anomalous scenes. Figure 6 gives a visualization of the anomalies caught by standard RNNs and SiTGRU. Each row consists of a sample anomalous frame (anomaly enclosed in red rectangle) and the equivalent residual heatmaps generated after evaluating different models with the anomalous frame.

Figure 7 presents regularity score and multiple anomaly visualization of our proposed method. Figure 7(a) shows regularity score of sample test videos from CUHK Avenue (video #6), UCSD Ped1 (video #1) and Ped2 (video #2), respectively. The regions highlighted in light-red show the presence of anomalous events (viz., person walking towards camera in Avenue, and biking in Ped1 and Ped2 datasets) in frames within shaded range. The figure also presents sample pairs of anomalous events and their equivalent heat maps from Avenue, Ped1 and Ped2 test sets ((b) - (d)) respectively. As shown in Figure 7(c) and (d), multiple abnormalities appearing on a single frame are identified and detected by the proposed structure.

| Optimizer | Loss | |
|---|---|---|
| | Mean squared error | Sparse categorical cross-entropy |
| AdaGrad | 0.702/0.353 | 0.682/0.367 |
| Adam | **0.718/0.342** | 0.698/0.350 |
| RMSprop | 0.655/0.389 | 0.621/0.401 |

Table 2: Performance comparison of models by changing loss functions and optimizers on Avenue test set.

| Optimizer | Loss | |
|---|---|---|
| | Mean squared error | Sparse categorical cross-entropy |
| AdaGrad | 0.685/0.354 | 0.691/0.351 |
| Adam | 0.706/0.333 | **0.731/0.321** |
| RMSprop | 0.676/0.359 | 0.692/0.343 |

Table 3: Performance comparison of models by changing loss functions and optimizers on Ped1 test set.

| Optimizer | Loss | |
|---|---|---|
| | Mean squared error | Sparse categorical cross-entropy |
| AdaGrad | 0.831/0.236 | 0.851/0.218 |
| Adam | 0.841/0.227 | **0.862/0.198** |
| RMSprop | 0.823/0.245 | 0.838/0.231 |

Table 4: Performance comparison of models by changing loss functions and optimizers on Ped2 test set.

### 4.4. Regression Analysis
#### 4.4.1. Loss-Optimizer Comparison

In addition to suppressing the involvement of the reset gate, the performance of our proposed method is affected by the loss function and optimizer we deploy for model learning. To assess the impact of these hyper-parameters, we represent the AUC/EER accuracy evaluation metrics as $AR$, and define a regression function $f$ that maximizes the overall value of $AR$ based on the type of loss function $L$ and optimizer $O$ we use:

$$AR(D_T) \approx f(D_T, L : O), \quad (9)$$

where $D_T = \{Avenue, Ped1, Ped2\}$. For a dataset $D_T$, the function $f$ is supposed to obtain a feasible loss-optimizer pair $(L, O)$ that generates a model capable of yielding the best $AR$ value.

While training on all datasets, we learn our model using different loss functions and optimizers. We conduct our experiment using mean squared error and sparse categorical cross-entropy losses, and optimize our model with AdaGrad, Adam and RMSprop [36]. Tables 2 to 4 show the loss-optimizer correlation of models trained by changing the loss functions and optimizers, and the performance results obtained on Avenue, Ped1 and Ped2 test sets, respectively. We can observe that mean squared error objective function combined with Adam optimizer can better detect abnormalities on Avenue dataset (Table 2), while combining sparse categorical cross-entropy loss with Adam optimizer generates effective models for discriminating anomalies on UCSD Ped1 and Ped2 datasets (Table 3 and 4 respectively).

| Update gate | Reset gate | Training curve | Remark |
|---|---|---|---|
| ON | ON | Normal | GRU |
| ON | OFF | Normal | SiTGRU |
| OFF | ON | Not normal | GRU w/o update gate |
| OFF | OFF | Normal | RNN |

Table 5: Learning outcome by changing the state of update gate and reset gate.

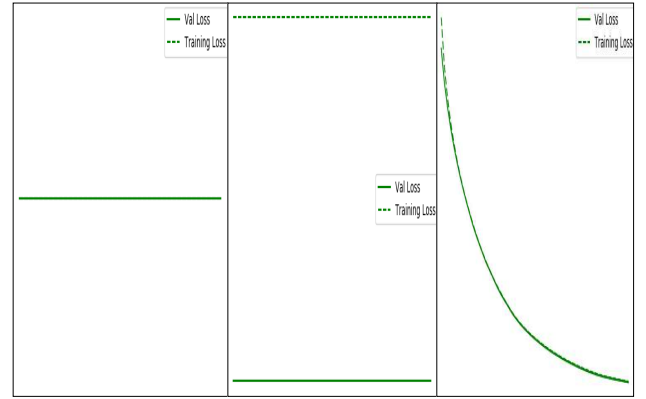

Figure 8: Training graph of GRU-based networks without update gate (left and middle) and SiTGRU (right).

#### 4.4.2. Reset-update gate Comparison

To further support the hypothesis that discarding the reset gate from GRUs can enhance performance effectiveness for abnormality detection, we carry out learning-behaviour-based categorical analysis on reset and update gates. We investigate the same network architecture by keeping the reset gate and removing the update gate from GRU cells. This changes the formulation of standard GRUs to the following form:

$$r_t = \delta(W_r x_t + U_r h_{t-1} + b_r) \quad (10a)$$

$$\tilde{h}_t = \tanh(W_h x_t + U_h(h_{t-1} \otimes r_t) + b_h) \quad (10b)$$

$$h_t = h_{t-1} \quad (10c)$$

As shown in Eq. (10c), the hidden state at current time step $h_t$ is not updating itself, which ignores the candidate memory state $\tilde{h}_t$ in Eq. (10b). In this case, the network is not able to learn spatio-temporal details across a sequence assigning the same content for every time step. This leads to a poorly learnt model with constant and insignificant loss values being produced during training. The models generated in this way are not capable of detecting anomalies well during evaluation as the normality learning is constrained due to the deployment of the sole reset gate. Table 5 highlights the impact of changing the status of the update and reset gates (turning *ON* and *OFF*), and shows the behaviour of the associated learning curves.

| Method | Ped1 | | Ped2 | |
|---|---|---|---|---|
| | AUC | EER | AUC | EER |
| GRU | 0.601 | 0.428 | 0.709 | 0.342 |
| LSTM | 0.600 | 0.432 | 0.701 | 0.353 |
| SiTGRU w/ ReLU | 0.624 | 0.393 | 0.723 | 0.322 |
| **SiTGRU** | **0.644** | **0.384** | **0.727** | **0.315** |

Table 6: Avenue model generalizability evaluation on Ped1 and Ped2 test sets.

| Method | Avenue | | Ped2 | |
|---|---|---|---|---|
| | AUC | EER | AUC | EER |
| GRU | 0.676 | 0.367 | 0.845 | 0.217 |
| LSTM | 0.664 | 0.352 | 0.823 | 0.241 |
| SiTGRU w/ ReLU | **0.690** | 0.360 | **0.872** | 0.210 |
| **SiTGRU** | 0.681 | **0.340** | **0.872** | **0.196** |

Table 7: Ped1 model generalizability evaluation on Avenue and Ped2 test sets.

| Method | Avenue | | Ped1 | |
|---|---|---|---|---|
| | AUC | EER | AUC | EER |
| GRU | 0.643 | 0.396 | 0.706 | 0.321 |
| LSTM | 0.624 | 0.418 | 0.684 | 0.336 |
| SiTGRU w/ ReLU | 0.646 | 0.405 | 0.586 | 0.419 |
| **SiTGRU** | **0.710** | **0.322** | **0.732** | **0.315** |

Table 8: Ped2 model generalizability evaluation on Avenue and Ped1 test sets.

Figure 8 shows the training graph generated by learning a network using the setups presented in Table 5. The graphs on the left and in the middle are generated by training a GRU without (w/o) update gate based network with different objective functions while the learning curve on the right is a result of training the proposed SiTGRU. The graph on the left shows that training and validation losses (solid green line) overlap linearly at the value of zero, and in the graph in the middle training and validation losses are still constant though not overlapping. The right most graph shows a normal learning curve where loss values decrease across epochs.

This indicates that the reset gate cannot stand on its own to learn the spatial and temporal nature of sequential data. Conversely, we can notice that the update gate is capable of replicating the function of the reset gate. Thus, discarding participation of the reset gate in GRU-based networks for video anomaly detection is a viable solution to enhance computational effectiveness and efficiency of models.

### 4.5. Generalization Study

We conduct different model evaluations to assess the generalizing capability of our trained models on unseen environments. We perform various sets of experiments to evaluate if a model trained on a particular dataset is able to effectively identify and detect anomalies when evaluated on another dataset. For each dataset, we pick the model that yields the best performance result in Table 1, and evaluate it on a dataset from a different category. Tables 6 to 8 depict generalization performance evaluation of models trained on a particular dataset and tested with video frames from another dataset.

We also evaluate the generalizing capability of our proposed model by carrying out model training on a new dataset prepared by merging randomly selected video samples from Avenue, Ped1 and Ped2 train sets. We select half of the videos from each dataset's train set and build a new train set with 66 videos. In addition, we also prepare a new test set of 69 videos by merging all test sets from Avenue, Ped1 and Ped2 datasets. Table 9 summarizes details of the videos that we select from each dataset's train set and the merged test set. We then train the proposed network on the train set of the new dataset and evaluate the trained model with the merged test set. Table 10 highlights generalization performance of the model with this scenario showing better evaluation achievement of our SiTGRU than standard RNNs.

We can observe from these tables that the proposed optimized-GRU-based deep model outperforms standard variants of GRU and LSTM for abnormality generalization task. In addition, our sigmoid based, optimized-GRU model discriminates abnormalities better on unrelated datasets than our ReLU based architecture.

### 4.6. Detecting Motion Anomalies

In the previous sections, we have shown the performance achievements and improvements gained by our proposed scheme for abnormality detection and generalization task. In addition to these sets of experiments, we further split the abnormality detection task to focus only on the detection of motion anomalies. Specifically, we choose the Avenue dataset for this task as it has videos that contain only motion abnormalities in its test set. Following the approach presented by Hinami et al. [17], we discard five videos which contain static appearance anomalies (i.e., video#1, video#2, video#8, video#9, video#10) from the test set and keep the rest sixteen videos. We call this test set *Avenue16* and evaluate our trained model with it. Experimental results in Table 11 show that our SiTGRU based model performs way better than standard RNN based models on this new test set. This shows that SiTGRU based models can identify and detect motion abnormalities better than standard RNNs implying their flexibility to adjust themselves to different types of anomalies.

### 4.7. Computational Efficiency

In addition to its improved abnormality detection performance on benchmark datasets, the other major feature of our SiTGRU framework proposed in this work is the enhanced computational efficiency. Figure 9 elaborates the maximum and minimum training time analysis of SiTGRU

Single-Tunnelled GRU for Abnormality Detection

| Type | CUHK Avenue | UCSD Ped1 | UCSD Ped2 | Total videos |
|---|---|---|---|---|
| Number of training videos | 16 | 34 | 16 | 66 |
| Number of selected videos | 8 | 17 | 8 | 33 |
| Selected videos | #1 #3 #5 #7 #9 #11 #13 #15 | #2 #4 #6 #8 #10 #12 #14 #16 #18 #20 #22 #24 #26 #28 #30 #32 #34 | #1 #3 #5 #7 #9 #11 #13 #15 | 33 |
| Number of testing videos (merged set) | 21 | 36 | 12 | 69 |

Table 9: Summary of merged test set and selected videos from the train sets of Avenue, Ped1 and Ped2 for generalizability assessment.

| Method | Merged dataset | |
|---|---|---|
| | AUC | EER |
| GRU | 0.634 | 0.388 |
| LSTM | 0.625 | 0.392 |
| **SiTGRU** | **0.707** | **0.343** |

Table 10: Generalization performance comparison with standard RNNs on a merged test set.

| Method | Avenue16 | |
|---|---|---|
| | AUC | EER |
| GRU | 0.563 | 0.425 |
| LSTM | 0.583 | 0.438 |
| **SiTGRU** | **0.744** | **0.305** |

Table 11: Performance comparison with standard RNNs on Avenue16 test set.

| Architecture | Avenue | Ped1 | Ped2 |
|---|---|---|---|
| GRU | 3941/3361 | 1522/1045 | 753/495 |
| LSTM | 4650/4376 | 1751/1307 | 659/648 |
| **SiTGRU** | **2140/2104** | **1268/832** | **432/308** |

Table 12: Summary of maximum-by-minimum per-epoch training time (sec) taken by standard GRU, standard LSTM, and SiTGRU on AED datasets.

| Method | Dataset | | |
|---|---|---|---|
| | Avenue | Ped1 | Ped2 |
| GRU | 10 | 7.2 | 8.2 |
| LSTM | 9.2 | 6.7 | 7.9 |
| **SiTGRU** | **11** | **7.8** | **9.2** |

Table 13: Comparison of test efficiency (fps) with standard RNNs.

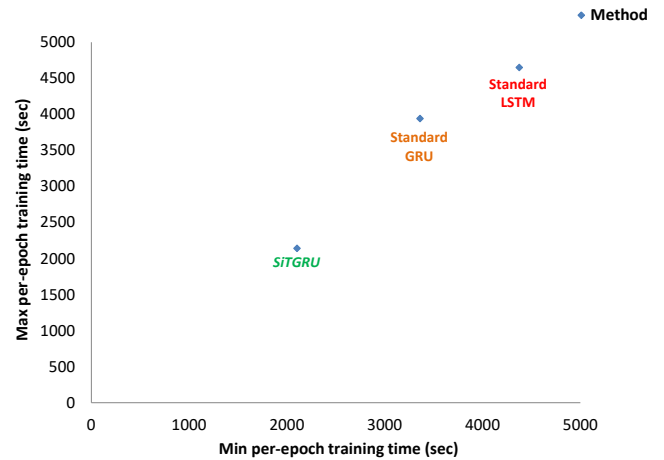

Figure 9: Comparison of different methods training efficiency on Avenue dataset.

and standard RNNs on Avenue dataset. Figures 10 and 11 give similar analyses for training on Ped1 and Ped2 datasets, respectively. Table 12 outlines the maximum and minimum per-epoch training time elapsed to generate models with similarly-structured standard GRU, standard LSTM and SiTGRU. Our SiTGRU takes less training time across all datasets which is basically attributed to the absence of the reset gate.

The same computational efficiency is also manifested during testing process where SiTGRU trained models take lesser testing time than standard recurrent networks (shown in Table 13). We perform all training and testing on a single NVIDIA GeForce GTX 1080 GPU on Keras [7] with Tensorflow [1] backend. Such computationally inexpensive and optimized models can easily be made accessible for deployment to evaluate abnormality detection on portable devices.

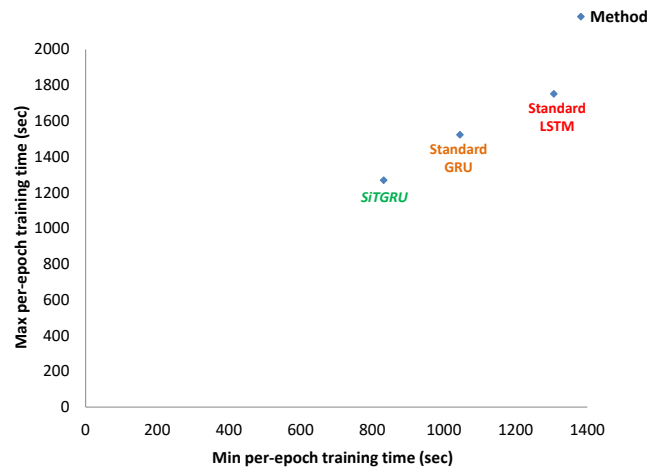

Figure 10: Comparison of different methods training efficiency on Ped1 dataset.

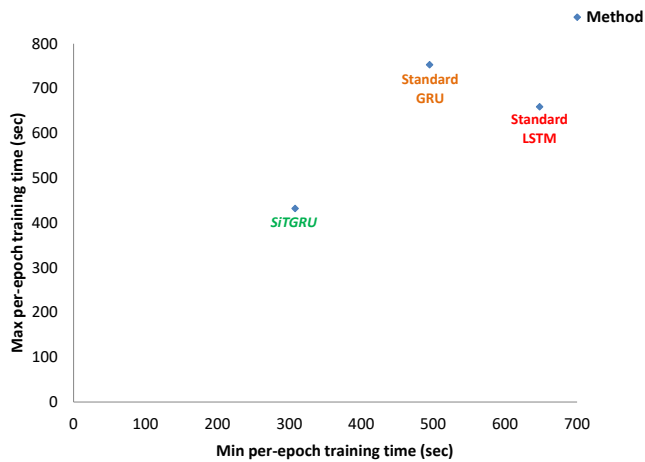

Figure 11: Comparison of different methods training efficiency on Ped2 dataset.

## 5. Conclusion

In this paper, we present an end-to-end Single-Tunnelled GRU based approach for abnormality detection and generalization in videos. We propose a novel encoding-decoding deep architecture that stacks GRU layers of varying number of units. We modify the standard GRU by leaving the reset gate from the cell structure and replacing the hyperbolic tangent activation with sigmoid activation. The experiments we carry out on CUHK Avenue and UCSD benchmarks signify the effectiveness of the proposed model for both detection and generalization tasks obtaining better performance than standard recurrent networks. Computational efficiency is also significantly improved resulting in lesser training and testing time, and reduced memory consumption.

Future works may investigate improving the introduced model by fusing with other variants of recurrent and deep networks. Another direction may involve investigating such model's deployment for pixel-level abnormality detection. Generalizing detection models across different datasets is still a challenging task that also needs further attention in future endeavours.

## Acknowledgments

This work is supported by the National Natural Science Foundation of China (No. 61972157), the Science and Technology Commission of Shanghai Municipality (No. 18D1205903), the Science and Technology Commission of Pudong (No. PKJ2018-Y46), and the Multidisciplinary Project of Shanghai Jiao Tong University (No. ZH2018ZDA25). It is also partially supported by the joint project of Tencent YouTu and Shanghai Jiao Tong University.

## Declaration of Competing Interest

None.

<i>Single-Tunnelled GRU for Abnormality Detection</i>

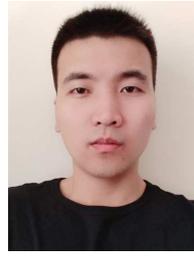

**Zhiwen Shao** received his B.Eng. degree in Computer Science and Technology from the Northwestern Polytechnical University, China in 2015. He will receive the Ph.D. degree from the Shanghai Jiao Tong University, China in 2020. After that, he will be an Assistant Professor at the School of Computer Science and Engineering, Southeast University, China. From 2017 to 2018, he was a joint Ph.D. student at the Multimedia and Interactive Computing Lab, Nanyang Technological University, Singapore. His research interests lie in face analysis and deep learning, in particular, facial expression recognition, facial expression manipulation, and face alignment.

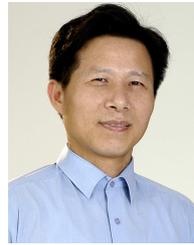

**Lizhuang Ma** received his B.S. and Ph.D. degrees from the Zhejiang University, China in 1985 and 1991, respectively. He is now a Distinguished Professor, Ph.D. Tutor, and the Head of the Digital Media and Computer Vision Laboratory at the Department of Computer Science and Engineering, Shanghai Jiao Tong University, China. He was a Visiting Professor at the Frounhofer IGD, Darmstadt, Germany in 1998, and was a Visiting Professor at the Center for Advanced Media Technology, Nanyang Technological University, Singapore from 1999 to 2000. He has published more than 200 academic research papers in both domestic and international journals. His research interests include computer aided geometric design, computer graphics, scientific data visualization, computer animation, digital media technology, and theory and applications for computer graphics, CAD/CAM.

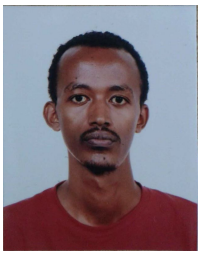

**Habtamu Fanta** received the B.S. Degree in Computer Science from the Mekelle University, Ethiopia in 2006, and the M.S. Degree in Information Science from the Addis Ababa University, Ethiopia in 2010. He is now a Ph.D. candidate in the Department of Computer Science and Engineering, Shanghai Jiao Tong University, China. His current research interests include abnormal event detection (AED) and application of deep recurrent neural networks for AED.